\documentclass[letterpaper, 10 pt, conference]{ieeeconf}  

\IEEEoverridecommandlockouts                              

\usepackage{graphics} 
\usepackage{epsfig} 
\usepackage{mathptmx} 
\usepackage{times} 
\usepackage{amsmath} 
\usepackage{amssymb}  
\usepackage{amsfonts}
\usepackage{algorithm}
\usepackage{algpseudocode}
\usepackage{booktabs}
\usepackage{caption}
\usepackage{subcaption}

\usepackage{etoolbox}
\usepackage{tikz}
\usepackage{booktabs}
\usepackage{tabularx}
\usepackage{array}
\usepackage{flushend}
\usepackage{multirow}
\usepackage{tcolorbox}
\usepackage{hyperref}
\usepackage{cite}

\newcolumntype{Y}{>{\centering\arraybackslash}p{2.5cm}}  
\newcolumntype{Z}{>{\centering\arraybackslash}p{2cm}}  
\newcolumntype{M}{>{\centering\arraybackslash}p{1.8cm}}  
\newcolumntype{N}{>{\centering\arraybackslash}p{1.5cm}} 
\newcolumntype{O}{>{\centering\arraybackslash}p{4.0cm}} 

\newcommand{\tikzsquare}[2][red,fill=red]{\tikz[baseline=-0.1ex]\draw[#1] (0,0) rectangle ++(#2,#2);}

\newcommand{\tikzcircle}[2][red,fill=red]{\tikz[baseline=-0.5ex]\draw[#1,radius=#2] (0,0) circle ;}%
\makeatletter
\patchcmd{\@makecaption}
  {\scshape}
  {}
  {}
  {}
\makeatletter
\patchcmd{\@makecaption}
  {\\}
  {.\ }
  {}
  {}
\makeatother

\usetikzlibrary{tikzmark}
\newcommand{\MapGenerator}{%
    \tikzmarknode[fill={rgb,255:red,244;green,142;blue,127},fill opacity=0.8,draw=black,inner sep=2pt,text opacity=1]{test}{Map Generator}%
}
\newcommand{\EnvUpdater}{%
    \tikzmarknode[fill={rgb,255:red,141;green,232;blue,239},fill opacity=0.8,draw=black,inner sep=2pt,text opacity=1]{test}{Environment Updater}%
}
\newcommand{\PercModule}{%
    \tikzmarknode[fill={rgb,255:red,141;green,232;blue,239},fill opacity=0.8,draw=black,inner sep=2pt,text opacity=1]{test}{Perception Module}%
}
\newcommand{\PlanModule}{%
    \tikzmarknode[fill={rgb,255:red,141;green,232;blue,239},fill opacity=0.8,draw=black,inner sep=2pt,text opacity=1]{test}{Planning Module}%
}
\newcommand{\ExpRecorder}{%
    \tikzmarknode[fill={rgb,255:red,249;green,143;blue,37},fill opacity=0.8,draw=black,inner sep=2pt,text opacity=1]{test}{Experiment Recorder}%
}
\newcommand{\MetricCalc}{%
    \tikzmarknode[fill={rgb,255:red,130;green,238;blue,198},fill opacity=0.8,draw=black,inner sep=2pt,text opacity=1]{test}{Metric Calculator}%
}
\newcommand{\MetricAnlz}{%
    \tikzmarknode[fill={rgb,255:red,130;green,238;blue,198},fill opacity=0.8,draw=black,inner sep=2pt,text opacity=1]{test}{Metric Analyzer}%
}

\title{\LARGE \bf
Evaluating Dynamic Environment Difficulty for Obstacle Avoidance Benchmarking
}

\author{Moji Shi, Gang Chen, Álvaro Serra Gómez, Siyuan Wu and Javier Alonso-Mora
\thanks{The authors are with the Department of Cognitive Robotics (CoR), Delft
University of Technology, 2628CD Delft, The Netherlands
{\tt\small \{m.shi-5; s.wu-14\}@student.tudelft.nl; \{g.chen-5; A.SerraGomez; j.alonsomora\}@tudelft.nl}
}%
\thanks{
This work is funded in part by the European Union (ERC, INTERACT, 101041863). Views and opinions expressed are however those of the author(s) only and do not necessarily reflect those of the European Union or the European Research Council Executive Agency. Neither the European Union nor the granting authority can be held responsible for them.
}
}

\begin{document}
\bstctlcite{IEEEexample:BSTcontrol}

\maketitle
\thispagestyle{empty}
\pagestyle{empty}

\begin{abstract}
Dynamic obstacle avoidance is a popular research topic for autonomous systems, such as micro aerial vehicles and service robots. Accurately evaluating the performance of dynamic obstacle avoidance methods necessitates the establishment of a metric to quantify the environment's difficulty, a crucial aspect that remains unexplored.
In this paper, we propose four metrics 
to measure the difficulty of dynamic environments. These metrics aim to comprehensively capture the influence of obstacles' number, size, velocity, and other factors on the difficulty.
We compare the proposed metrics with existing static environment difficulty metrics and validate them through over 1.5 million trials in a customized simulator. This simulator excludes the effects of perception and control errors and supports different motion and gaze planners for obstacle avoidance.
The results indicate that the survivability metric outperforms and establishes a monotonic relationship between the success rate, with a Spearman's Rank Correlation Coefficient (SRCC) of over 0.9. Specifically, for every planner, lower survivability leads to a higher success rate.
This metric not only facilitates fair and comprehensive benchmarking but also provides insights for refining collision avoidance methods, thereby furthering the evolution of autonomous systems in dynamic environments.
\end{abstract}

\section{INTRODUCTION}

Dynamic obstacle avoidance is a popular research topic in the field of robotics  \cite{fineanWhereShouldLook2022,tordesillasPANTHERPerceptionAwareTrajectory2022a,liuSearchbasedMotionPlanning2017, muellerTrajGen, lopezAggressive3DCollision2017,chenActiveSenseAvoid2021}.
To evaluate the obstacle avoidance methods, most works handcraft their custom test maps in simulation or the real world to demonstrate a higher success rate. However, the difficulty of the chosen maps is not stated or evaluated. This indication of difficulty is vital. One method with a very high success rate in a simple environment may drastically fail in a more difficult environment. It is ideal to test the methods under different difficulty levels.
Furthermore, when comparing with other obstacle avoidance methods, pointing out the difficulty or creating environments with similar difficulties used in the baselines can promote a fair comparison.

\begin{figure}[!t]
    \centering
    \includegraphics[width=0.48\textwidth]{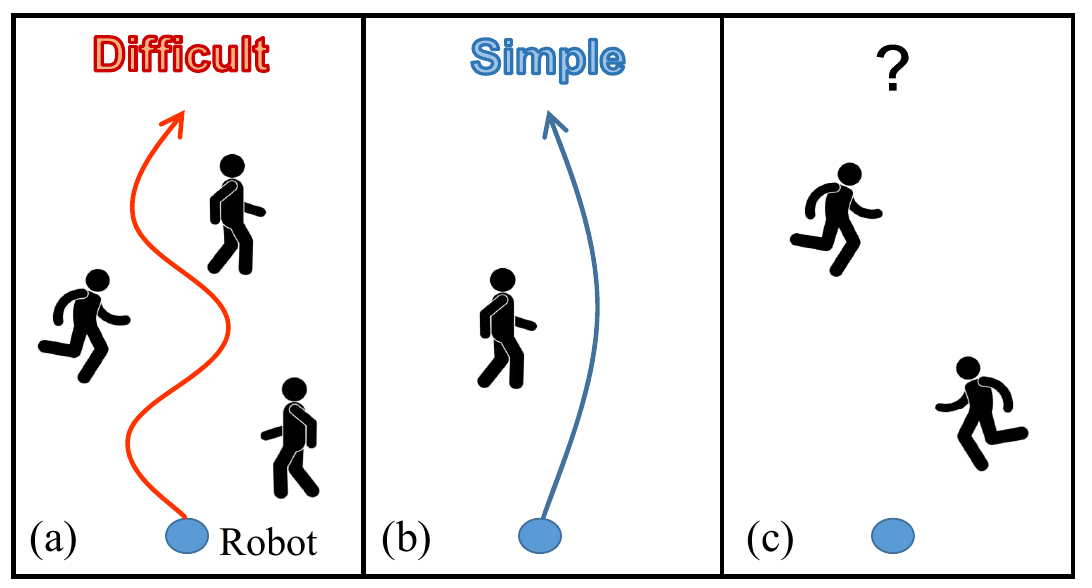}
    \caption{Dynamic environments with different difficulties. (a) shows a map with two walking pedestrians and one running pedestrian; (b) is with one walking pedestrian; (3) is with two running pedestrians. While it is intuitively clear that (b) represents a simpler environment than (a), determining whether (c) is simpler or more difficult than (a) remains unclear. The magnitude of the difficulty is hard to determine.}
    \label{fig:cover}
\end{figure}

In static environments, the difficulty is usually evaluated by the density of the environment \cite{ahmad2021autonomous,borenstein1990real}.
In dynamic environments, defining a difficulty metric is much more complicated because the difficulty is influenced by many factors \cite{nairDynaBARNBenchmarkingMetric2022}, 
such as obstacle size, number, velocity, and motion profile. While Fig \ref{fig:cover} makes it clear that Map (b) is less challenging than (a), discerning whether Map (c), with fewer but faster pedestrians, is more difficult than Map (a) remains a hard task without a quantitative metric. Currently, there is no existing metric to quantify the difficulty of a dynamic map.

In this paper, we design four quantitative metrics from different perspectives, such as survivability, traversability, and velocity obstacle (VO) feasibility, to evaluate the difficulty of a dynamic environment. Recognizing the comprehensive impact of various environmental factors on difficulty, our metrics avoid constructing a formula with these factors but try to capture the influence caused by them, starting from one single foundational premise: a map with higher difficulty should consistently result in a lower obstacle avoidance success rate. To evaluate the effectiveness of these metrics, we developed an efficient custom simulator that excludes the influence of perception and control error and computational power, while retaining the influence of environmental factors. Numerous tests using different motion and gaze planners are conducted to identify the best metric that aligns with our foundational premise. We then delve into a discussion of the merits and drawbacks of the metrics and present guidelines for their utilization in both simulation and real-world scenarios.




The code of using the metric and the simulator is available at \textbf{Code:} \href{https://github.com/smoggy-P/gym-Drone2D-ActivePerception}{https://github.com/smoggy-P/gym-Drone2D-ActivePerception} \textbf{Homepage:} \href{https://smoggy-p.github.io/Evaluating_Dynamic_Difficulty/}{https://smoggy-p.github.io/Evaluating\_Dynamic\_Difficulty/}

\section{Related Work}
\subsection{Collision avoidance in unknown dynamic environments}
Robot collision avoidance in dynamic, unpredictable environments is a longstanding challenge in robotics, even under full observability conditions. A plethora of solutions have emerged to tackle this issue. Methods in \cite{liuSearchbasedMotionPlanning2017, muellerTrajGen, lopezAggressive3DCollision2017,chenActiveSenseAvoid2021} utilize sampling-based methods to create multiple trajectories, from which the one of lowest cost is selected. In contrast, optimization-based methods like \cite{zhuChanceConstrainedCollisionAvoidance2019,mellingerMinimumSnapTrajectory2011,blackmoreChanceConstrainedOptimalPath2011} treat collision avoidance as an optimization problem, setting specific constraints to guide trajectory generation.

Gaze planning has proved to be important in dynamic obstacle avoidance. With real-time perception, dynamic obstacles require constant monitoring for future predictions. While researchers like \cite{chenOnlineGenerationCollisionfree2016,liuPlanningDynamicallyFeasible2017} discuss generating collision-free trajectories with 360 degrees of FOV, many real-world situations limit the range of FOV. This necessitates gaze planning to decide where the robot should look to gather more information for better collision avoidance. Simple strategies like those in \cite{limThreedimensional3DDynamic2019,chenRASTRiskAwareSpatioTemporal2023,tordesillasFASTERFastSafe2021} suggest looking in the velocity or target direction. More intricate methods, found in \cite{fineanWhereShouldLook2022,chenActiveSenseAvoid2021,zhouRAPTORRobustPerceptionaware2020,tordesillasPANTHERPerceptionAwareTrajectory2022a,tordesillasDeepPANTHERLearningBasedPerceptionAware2022}, treat gaze planning as an optimization problem, directing the gaze based on various objectives.

\subsection{Benchmarking for collision avoidance}
To further advance dynamic collision avoidance methods, it is essential to objectively compare them using benchmarks that evaluate performance and robustness across different conditions.
The most common benchmarking approach for collision avoidance methods involves comparing success rates across manually designed test environments \cite{fineanWhereShouldLook2022,tordesillasPANTHERPerceptionAwareTrajectory2022a,chenRASTRiskAwareSpatioTemporal2023}. Some studies have proposed more standardized benchmarking suites for collision avoidance in static environments, such as \cite{heidenBenchMRMotionPlanning2021,yu2023avoidbench,sturtevantBenchmarksGridBasedPathfinding2012,behzadan2019adversarial,singh2009steerbench}. These benchmarking suites provide randomly generated static maps and standardized evaluation pipeline to guarantee fair comparison. Furthermore, some metrics are defined to quantify the static environments difficulty like obstacle density \cite{ribeiro2005obstacle, lopez2017aggressive} and traversability  \cite{nousPerformanceEvaluationObstacle2016,perilleBenchmarkingMetricGround2020, wang2023curriculum}. These metrics provide more insights into the testing maps and thus allow a more comprehensive comparison between collision avoidance methods.

For benchmarking in dynamic collision avoidance, \cite{nairDynaBARNBenchmarkingMetric2022,kastnerArenaBenchBenchmarkingSuite2022,martinez2009benchmarking,martinez2009collision} provide standard benchmarking suites where collision avoidance methods can be tested in randomly generated dynamic maps and be compared. \cite{nairDynaBARNBenchmarkingMetric2022} also mentions the non-quantitative difficulty of dynamic maps. However, they lack a comprehensive set of metrics to quantify the difficulty of the dynamic environments and the validation of these metrics.

\section{SIMULATOR DESIGN}\label{sec:sim}

These metrics will be introduced in Sec. \ref{sec:metric}.  
In this section, the designed simulator is introduced to validate the proposed difficulty metrics presented in Sec. \ref{sec:metric}.

\subsection{Requirements and Assumptions}
\label{sec:assumption}
We validate the difficulty metrics by investigating their correlation with the performance of typical planners (i.e. How the success rate of planners changes when the difficulty metrics change). Thus, an ideal simulator should adhere to several criteria:

\begin{itemize}
\item Comprehensive and rapid assessment of different collision avoidance methods(including trajectory planners and gaze planners mentioned in Sec. \ref{sec:traj}).
\item Ensure the isolation of map difficulty as a variable affecting the performance of various methods, excluding the effect from factors such as perception error, control error, and computational power.
\end{itemize}

Meanwhile, several assumptions have been made to simplify the simulator: a) The environment is planar. b) Only disc-shaped dynamic obstacles are employed.

\subsection{Simulator Pipeline}
\begin{figure}[t!]
    \centering
    \begin{subfigure}[t]{0.155\textwidth}
        \centering
        \includegraphics[width=\textwidth]{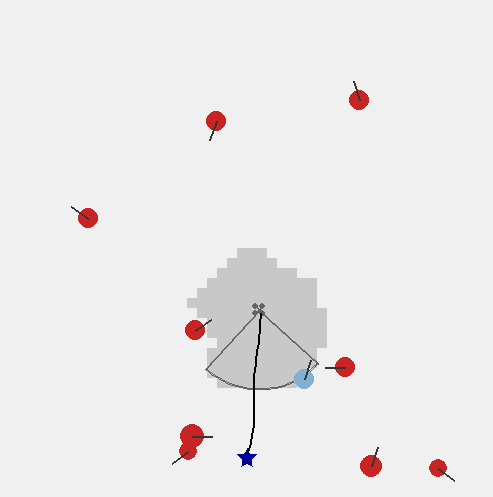}
    \end{subfigure}
    \begin{subfigure}[t]{0.155\textwidth}
        \centering
        \includegraphics[width=\textwidth]{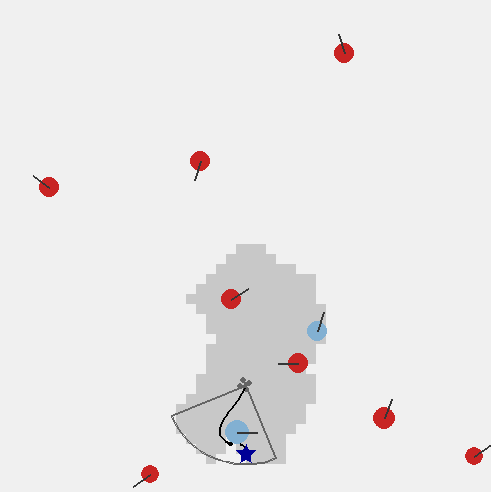}
    \end{subfigure}
    \begin{subfigure}[t]{0.155\textwidth}
        \centering
        \includegraphics[width=\textwidth]{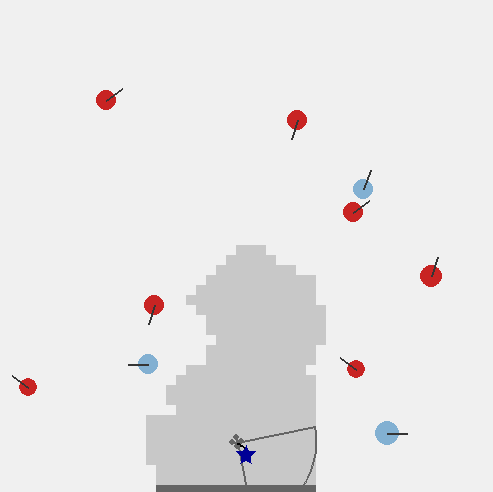}
    \end{subfigure}
\caption{Visualization of one experiment trial in the custom simulator. \tikzcircle[fill={rgb,255:red,200;green,36;blue,35}]{3pt}: Unobserved Obstacle; \tikzcircle[fill={rgb,255:red,130;green,176;blue,210}]{3pt}: Observed Obstacle; \tikzsquare[fill={rgb,255:red,240;green,240;blue,240}]{5pt}: Unexplored Area; \tikzsquare[fill=
{rgb,255:red,200;green,200;blue,200}]{5pt}: Unoccupied Area}
\label{fig: env}
\end{figure}

We design our simulator following the OpenAI gym standard, informed by insights from \cite{korber2021comparing} due to its lightweight feature and its “agent-environment” loop design. Based on that, we realize a sequential navigation pipeline that allows us to exclude the perception noise, control noise, and computational time. As shown in Fig. \ref{fig:pipeline}, our simulator is composed of the following components:

\begin{figure}[!t]
    \centering
    \includegraphics[width=1.0\linewidth]{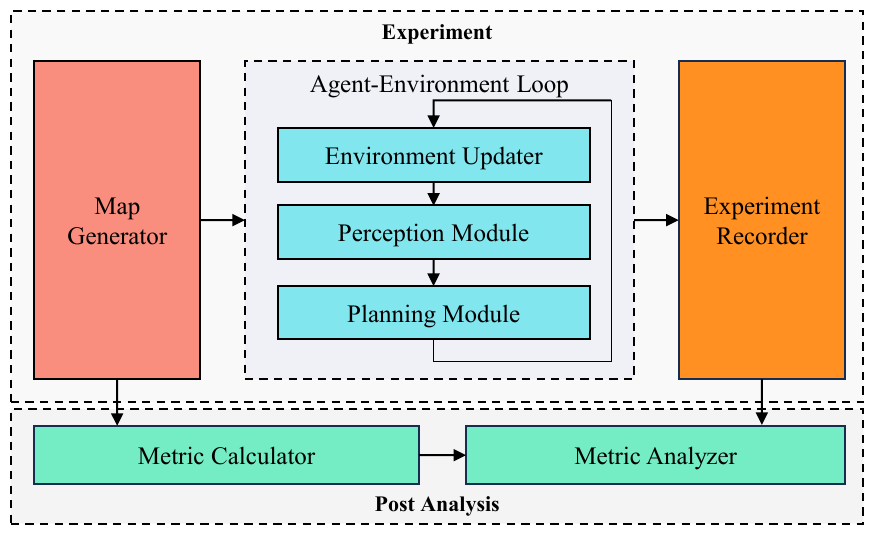}
    \caption{The pipeline of the proposed custom simulator}
    \label{fig:pipeline}
\end{figure}

a) \MapGenerator\label{sec:map_generator} creates random maps with varying dynamic obstacles in terms of number, size, velocity, and motion profiles. Motion profiles determine the movement of dynamic obstacles, further detailed in the Sec. \ref{sec:updater}. We craft two map datasets using this generator. \textit{Dataset I} assumes uniform size and velocity for all obstacles in a map. The map is thus characterized by the number of obstacles $n_{obs}$, obstacles' size $r_{obs}$, and velocity $v_{obs}$. 
A random seed is used to initialize the obstacles' position and direction.
In contrast, \textit{Dataset II} is for more general scenarios with varying obstacle sizes, velocities, and motion profiles. One example of the generated map can be found in Fig. \ref{fig: env}.

b) \EnvUpdater\label{sec:updater} updates dynamic obstacles' position according to the motion profile defined by Map Generator. In \textit{Dataset I}, the motion profile is defined as the constant velocity model (CVM). In \textit{Dataset II}, it can be defined as the reciprocal velocity obstacle (RVO) \cite{vandenbergReciprocalVelocityObstacles2008} which makes obstacles to avoid each other.

c) \PercModule\label{sec:perception} applies a ray-casting algorithm to simulate the FOV of the robot, producing a 2D occupancy map and dynamic trackers. Grids can be unexplored, unoccupied, or occupied. Observation of dynamic obstacles is updated using a Kalman Filter (KF) tracker, with visible grids and obstacles directly updated to their ground truth, eliminating perception noise. The perception output can also be seen in Fig. \ref{fig: env}.

d) \PlanModule\label{sec:planning} processes perception outputs to generate the future trajectory and yaw angle velocity using the planners to be evaluated in table \ref{tab:methods_comparison}. If no trajectory is feasible, a replan is triggered for the next step and outputs a braking command. The output trajectory is directly executed, ensuring no control errors impact planner performance.

e) \ExpRecorder\label{sec:recorder} logs results, categorizing outcomes as Success, Collision, or Deadlock. Deadlock arises when the robot fails to find a feasible trajectory after 5 consecutive replan attempts.

After achieving the experiment results, we calculate the metrics of maps and analyze the correlation with the success rates through \MetricCalc and \MetricAnlz. Details of these two modules will be explained in Sec. \ref{sec:metric}

\section{METRICS DESIGN}\label {sec:metric}
In this section, metrics are defined to quantify dynamic environment difficulty corresponding to \MetricCalc~ in Fig. \ref{fig:pipeline} including Obstacle Density \cite{ahmad2021autonomous}, Traversability \cite{yu2023avoidbench, nousPerformanceEvaluationObstacle2016}, Dynamic Traversability, VO Feasibility, Survivability, and Global Survivability. The expected correlation between these metrics and map difficulty is shown in Tab. \ref{tab:metric_correlation}. Obstacle Density and Traversability are two existing difficulty metrics for static environments and are introduced as baselines. The other four metrics are designed by us to measure dynamic environment difficulty. Finally, we detail the evaluation method for the metrics.
\bgroup
\def\arraystretch{1.2}
\begin{table}[!h]
    \centering
    \begin{tabularx}{0.49\textwidth}{@{}cMNO@{}}
    \toprule
    & \textbf{Metric} & \textbf{Expected Correlation with Difficulty} & \textbf{Description} \\ 
    \midrule
    1)& Obstacle Density \cite{ahmad2021autonomous} & $+$ & Maps with large obstacle density are considered difficult.  \\
    2)& Traversability \cite{yu2023avoidbench, nousPerformanceEvaluationObstacle2016} & $-$ & Maps with large traversable space are considered simple. \\
    3)& Dynamic Traversability & $-$ & Maps with large dynamic traversable space are considered simple. \\
    4)& VO Feasibility & $-$ & Maps with more feasible velocities are considered simple.  \\
    5)& Survivability & $-$ & Long survival time of sampled position means the map is simple. \\
    6)& Global Survivability & $-$ & Same as survivability. \\
    \bottomrule
    \end{tabularx}                             
    \caption{\label{tab:metric_correlation}Expected correlation between metrics, map difficulty, and success rate of planner. 
    We reverse some metrics to make sure all metrics increase monotonously as difficulty increases. Reversed metrics include Traversability, Dynamic Traversability, VO Feasibility, Survivability, and Global Survivability. This preprocessing step will be further introduced in Sec. \ref{sec:experiment_results}.}
\end{table}
\bgroup
\def\arraystretch{1}

\subsection{Difficulty Metrics}
\subsubsection{Obstacle Density}
Obstacle density is a widely used metric to quantify the difficulty of static environments \cite{ahmad2021autonomous,borenstein1990real}. It is defined as the areas occupied by obstacles divided by the total area of the map:
\begin{equation}
    \text{Obstacle Density}=\frac{A_{obs}}{A_{map}}
\end{equation}
In dynamic maps, obstacle density is irrelevant with time and thus will not change at different time steps as we suppose that obstacles do not overlap with each other. Thus, we only calculate it once when initializing the dynamic environment.

\subsubsection{Traversability}
Traversability is proposed in \cite{nousPerformanceEvaluationObstacle2016,yu2023avoidbench} to evaluate the difficulty of static environments. It is defined as the average traversable distance for all uniformly sampled positions and directions as shown in Fig. \ref{subfig:traversability}:
\begin{equation}
\text{Traversability}=\frac{1}{N}\sum_{i=1}^{N}d_{i}
\end{equation}
where $d_i$ represents the traversable distance at $i$-th sampled position and direction, and $N$ represents the total number of sampled positions and directions.

\subsubsection{Dynamic Traversability}
The traversability at different time steps might differ. We improve it for dynamic maps by designing dynamic traversability. The dynamic traversability is calculated by sampling the time step and averaging the traversability over the sampled time step. 

\begin{equation}
    \text{Dynamic Traversability}=\frac{1}{MN}\sum_{j=1}^{M}\sum_{i=1}^{N}d_i(t_j)
\end{equation}
where the $d_i(t_j)$ is the traversable distance at $i$-th sampled position and direction and at the time step $t_j$. The time step is sampled uniformly from the beginning of the dynamic map: $t_j =(j-1)\cdot t_{\text{sample}}, \forall j \in \{1,2\cdots M\}$. $N$ is the number of sampled positions and directions, and $M$ is the number of sampled time steps.

\subsubsection{VO Feasibility}
 The velocity obstacle (VO) is introduced in \cite{fiorini1998motion} for multi-agent collision avoidance. The VO for robot A regarding a collision with obstacle B is given by:
    \begin{align}
        VO_{B} = \left\{\Vec{v_A}  | \exists t>0: (\Vec{v_A}-\Vec{v_B})t\in D(\Vec{p_B}-\Vec{p_A}, r_A+r_B)\right\}
    \end{align}
where $\Vec{v_A}$ and $\Vec{v_B}$ are the velocities of $A$ and $B$, $\Vec{p_A}$ and $\Vec{p_B}$ are the positions of $A$ and $B$, $r_A$ and $r_B$ are the radius of $A$ and $B$, and $D(\Vec{p_B}-\Vec{p_A}, r_A+r_B)$ is the disk centered at $\Vec{p_B}-\Vec{p_A}$ with radius $r_A+r_B$. The union of all VOs determines infeasible velocities for the ego-robot. Intuitively, a larger VO area implies a more challenging environment. Hence, we propose a VO feasibility metric, where we first sample $N$ positions around the map. At each position indexed by $i$, we sample $n_{vel}$ velocities and calculate the percentage of sampled velocities outside any VO:

\begin{align}
    \text{VO Feasibility} = \frac{1}{N}\sum_{i=1}^{N}\frac{n_{feasible}(i)}{n_{feasible}(i)+n_{infeasible}(i)}
\end{align}

$n_{feasible}$ and $n_{infeasible}$ denote the number of sampled velocities that lie outside and inside the VO area as shown in Fig. \ref{subfig:vo_feasibility} and $n_{vel}=n_{feasible}(i)+n_{infeasible}(i), \forall i\in\{1,2\cdots N\}$.

\subsubsection{Survivability}
\label{sec:survivability}
The aforementioned metrics mainly assess discrete difficulty as they are calculated at certain time steps. For example, Obstacle Density, Traversability, and VO feasibility are only calculated at the initial step of the dynamic map, and dynamic traversability samples $M$ discrete steps. These metrics neglect the continuous changes in the environment. To address this issue, the survivability metric is proposed. We assume that static robots are placed at sampled positions and calculate their average survival time. The survival time is defined as the duration from the initial time step until one obstacle moves into and collides with the static robot, as shown in Fig. \ref{subfig:survivability}:
    \begin{align}
        \text{Survivability} = \frac{1}{N}\sum_{i=1}^{N}\min (t_i, T_{max})
    \end{align}
where $t_i$ is the surviving time of the robot at the $i$-th sample, $N$ is the number of robot samples, and $T_{max}$ is the upper bound of the survival time for normalization. The $N$ positions are sampled from a uniform grid with $d_{sample}$ as the distance between two grid points. 

Note that in the real-world tests, there is no need to actually ``place'' static robots on the map. We can record the trajectories of all obstacles and calculate the Survivability by replaying these trajectories and recording survival time in these replays.

\subsubsection{Global Survivability}
Instead of putting one static robot on the map at each sample in the Survivability calculation, Global Survivability is calculated by assuming that $N$ robots are simultaneously placed at different positions across the map. For each deployment, the duration from time step $t_j$ until any robot collides with an obstacle is recorded. Global survivability quantifies this average survival time, considering deployments from multiple start time steps $t_j$:
    \begin{align}
        \text{Global Survivability} = \frac{1}{K}\sum_{j=1}^{K}\min(t_j^1, t_j^2 \dots, t_j^N, T_{max})
    \end{align}
where $t_j^n$ is the surviving time of the $n$-th robot starting from $t_j$ (i.e. The duration from $t_j$ until collision), and $K$ is the number of samples of start time steps.

\begin{figure}[t!]
    \centering
    \begin{subfigure}[t]{0.155\textwidth}
        \centering
        \includegraphics[width=\textwidth]{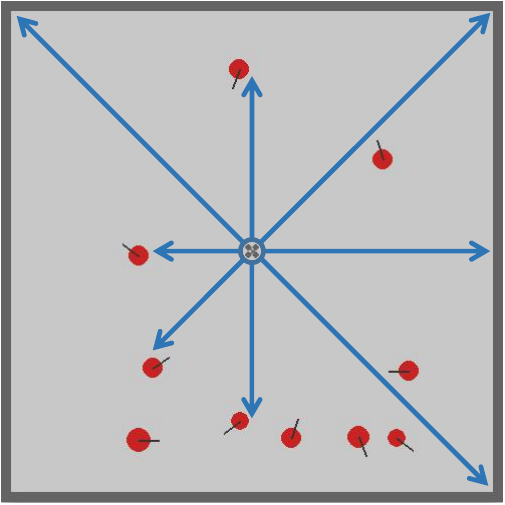}
        \caption{Traversability}
        \label{subfig:traversability}
    \end{subfigure}
    \begin{subfigure}[t]{0.155\textwidth}
        \centering
        \includegraphics[width=\textwidth]{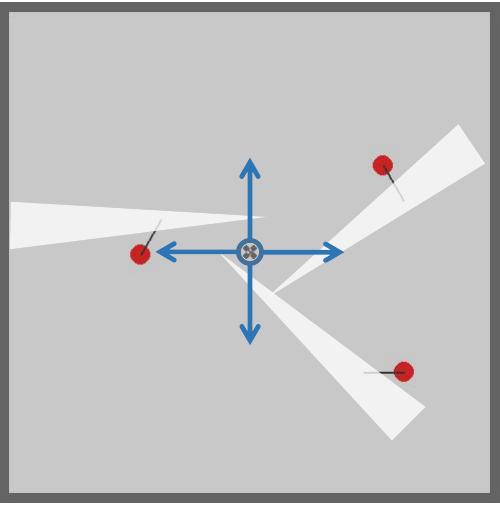}
        \caption{VO Feasibility}
        \label{subfig:vo_feasibility}
    \end{subfigure}
    \begin{subfigure}[t]{0.155\textwidth}
        \centering
        \includegraphics[width=\textwidth]{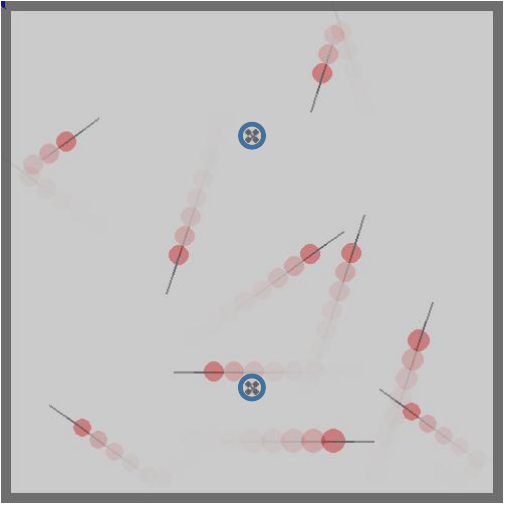}
        \caption{Survivability}
        \label{subfig:survivability}
    \end{subfigure}
\caption{Examples of metrics calculation. In (a), traversability is the average of traversable distances(shown by blue arrows) in 8 uniformly distributed directions at one sampled position. In (b), the white areas represent the infeasible VO regions. 4 velocity vectors(shown by blue arrows) are sampled at the sampled position. Only the velocity pointing right lies in the infeasible VO regions. So we have $n_{feasible}=3$ and $n_{infeasible}=1$. The resulting VO feasibility is thus $\frac{3}{4}$. In (c), a simple example of Survivability calculation is shown. The static robot is assumed to be placed in two positions. The static robot above does not intersect with any obstacle history trajectories in $T_{max}$ while the static robot below survives 2 seconds until it collides with the obstacle heading toward the left. The survivability is thus $(T_{max}+2)/2=(3+2)/2=2.5$}
\label{fig: metric_definition}
\end{figure}

\subsection{Evaluation Methodology}
\label{sec:eval_method}
We want to evaluate all metrics in \textit{Dataset I} mentioned in Sec. \ref{sec:map_generator}. Assuming \textit{Dataset I} contains $n$ different maps, each denoted by $M_i$. The dataset can be denoted as $\mathcal{M} = \left\{M_1, M_2, ..., M_n\right\}$. $m$ different planners are tested on these maps. The map difficulty is defined as a scalar function $\text{D}(M_i)$. The success rate of a planner $j$ on map $M_i$ is defined as $SR_i^j$. We introduce two quantitative indicators for evaluating whether the metrics are representative of the dynamic map difficulty:

\textbf{Spearman's Rank Correlation Coefficient (SRCC):} This metric evaluates the monotonic relationship between two variables. For a given metric \( \text{D}(M_i) \), we evaluate its monotonical relation with the success rate \( SR_{i}^j \) for planner \( j \) and denote it as $SRCC_j$. The overall effectiveness of the difficulty metric is the average of \( SRCC_j \).

\textbf{Coefficient of Variation (CV):} This metric suggests the variation of the success rate $SR_{i}^j$ on maps with close difficulty metrics. The smaller $CV$ value means that if two maps have very close metrics, the success rate of one planner on these two maps will not differ too much, indicating a good correlation between the metric and the success rate. Here, we round the metric to the nearest integers to construct map groups so that each group includes maps with similar metrics:
    \begin{align}
        \mathcal{M}_k &= \left\{M_i | \text{D}(M_i) \in [k, k+1]\right\}\ k=0,1,2,\dots,9
    \end{align}
For each group \( \mathcal{M}_k \), \( CV_{j}^k \) for planner \( j \) denotes performance stability:

\begin{align}
    CV_j^k=\frac{\sigma_j^k}{\mu_j^k}
\end{align}
$\sigma_j^k$ and $\mu_j^k$ are the standard deviation and the mean value of the success rate of planner $j$ in map group $\mathcal{M}_k$. The $CV$ measure for the difficulty metric is given by the average of \( CV_{j}^k \).

\section{EXPERIMENT}\label{sec:experiment}
To evaluate the metrics introduced in Sec. \ref{sec:metric}, we replicate various trajectory and gaze planners, testing them within two map datasets outlined in Sec. \ref{sec:map_generator}. After testing, we compute metrics for each map, aiming to explore the relationship between these metrics and planner success rates.

\subsection{Experiment Setup}
\subsubsection{Map Dataset}
All maps are generated in a $50m\times 50m$ square area. As mentioned in Sec.  \ref{sec:sim}, maps in \textit{Dataset I} are generated by defining 3 variables and one random seed. The range of these variables is shown in Tab.  \ref{tab: map dataset 1}. For each variable setting, we generate 20 maps with different random seeds. We will thus have $3\times 3\times 3\times 20=540$ maps in \textit{Dataset I}.

\begin{table}[!h]
    \centering
    \begin{tabular}{@{}cccc@{}}
    \toprule
    \textbf{Parameter} & $\mathbf{n_{obs}}$ & $\mathbf{r_{obs}}$           & $\mathbf{v_{obs}}$         \\ 
    \midrule
    \textbf{Range} & $\left\{10,20,30\right\}$ & $\left\{0.5,1,1.5\right\}m$ & $\left\{2, 4, 6\right\}m/s$ \\
    \bottomrule
    \end{tabular}
    \caption{\label{tab: map dataset 1}Range of variables for maps in \textit{Dataset I}}
\end{table}

If the metric performs well in \textit{Dataset I}, it is further evaluated in \textit{Dataset II}, where assumptions are relaxed and three more general map types are introduced: (a)  obstacle velocities within a map differ, sampled from the distribution $\left[2, 6\right]$ m/s; (b) obstacle sizes within a map differ, sampled from $\left[0.5, 1.5\right]$ m; (c) obstacles move using the RVO \cite{vandenbergReciprocalVelocityObstacles2008} motion profile. For each type, we generate 40 maps. Therefore, we have 120 maps in \textit{Dataset II}.

\subsubsection{Robot Parameters}
We assume that the robot is a 2D circle with a radius of 1m. The maximum acceleration of the robot is 4m/s$^2$. The depth of the FOV is 8m, and the range of the FOV can be 90 to simulate a single depth camera or 360 degrees to simulate multiple depth cameras and LiDAR. Since a larger yaw angle velocity will bring large errors in depth estimation, the maximum yaw angle velocity is set to 1.4 rad/s according to \cite{chenActiveSenseAvoid2021}.


\subsubsection{Planners}\label{sec:traj}
We use multiple planners for trajectory and gaze planner in Tab.  \ref{tab:methods_comparison} and Tab. \ref{tab:gaze_planning_methods} to ensure that the observed relationship remains consistent rather than being specific to a particular planner.

\bgroup
\def\arraystretch{1.5}
\begin{table}[!h]
    
    \begin{tabularx}{0.49\textwidth}{ZX}
    \toprule
    \textbf{Method} & \multicolumn{1}{c}{\textbf{Description}} \\
    \midrule
    Global Motion Primitive \cite{liuSearchbasedMotionPlanning2017} & Samples trajectories, filters these trajectories with collision checking and conducts a graph search in these trajectories until reaching the goal. \\ 
    Model Predictive Control (MPC) \cite{zhuChanceConstrainedCollisionAvoidance2019} & Defines the cost as the distance between future trajectory and target position plus control input, and defines the collision constraints as the distance between the robot and obstacles which are modeled as ellipsoids. \\ 
    Local Primitive \cite{chenActiveSenseAvoid2021} & Samples local targets, generates an optimal local trajectory \cite{muellerTrajGen} towards the local target with the lowest cost, and iteratively updates the local target until reaching the global target. \\
    \bottomrule
    \end{tabularx}
    \caption{Baselines of Trajectory Planner}
    \label{tab:methods_comparison}
\end{table}

Here the trajectory planners are chosen in terms of different planning horizons (\cite{liuSearchbasedMotionPlanning2017} generates a global trajectory while \cite{zhuChanceConstrainedCollisionAvoidance2019, chenActiveSenseAvoid2021} generate local trajectory) and planning methods(\cite{liuSearchbasedMotionPlanning2017, chenActiveSenseAvoid2021} are sampling-based methods while \cite{zhuChanceConstrainedCollisionAvoidance2019} is optimization-based method).

\begin{table}[ht]
    \centering
    \begin{tabularx}{0.48\textwidth}{ZX}
        \toprule
        \textbf{Method} & \multicolumn{1}{c}{\textbf{Description}} \\
        \midrule
        FullRange & The perception range is expanded to 360 degrees. No gaze planner is needed in this case. \\ 
        LookAhead \cite{limThreedimensional3DDynamic2019} &Look at the current velocity direction. \\
        LookGoal \cite{tordesillasFASTERFastSafe2021} & Look at the current target direction. \\
        Rotating & Rotate in the largest rotating velocity\\
        Finean et al. \cite{fineanWhereShouldLook2022} & Optimize an objective function to find a trade-off between looking at the future trajectory and looking at grids that have not been updated for a long time. \\
        Owl \cite{chenActiveSenseAvoid2021} & Optimize multiple objectives, including looking at the velocity direction, the target position, the direction that has not been updated for a long time, and the observed dynamic obstacles. \\
        \bottomrule
    \end{tabularx}
    \caption{Baselines of Gaze Planner}
    \label{tab:gaze_planning_methods}
\end{table}

\bgroup
\def\arraystretch{1}%
\begin{figure*}[t]
    \centering
    \includegraphics[width=\textwidth]{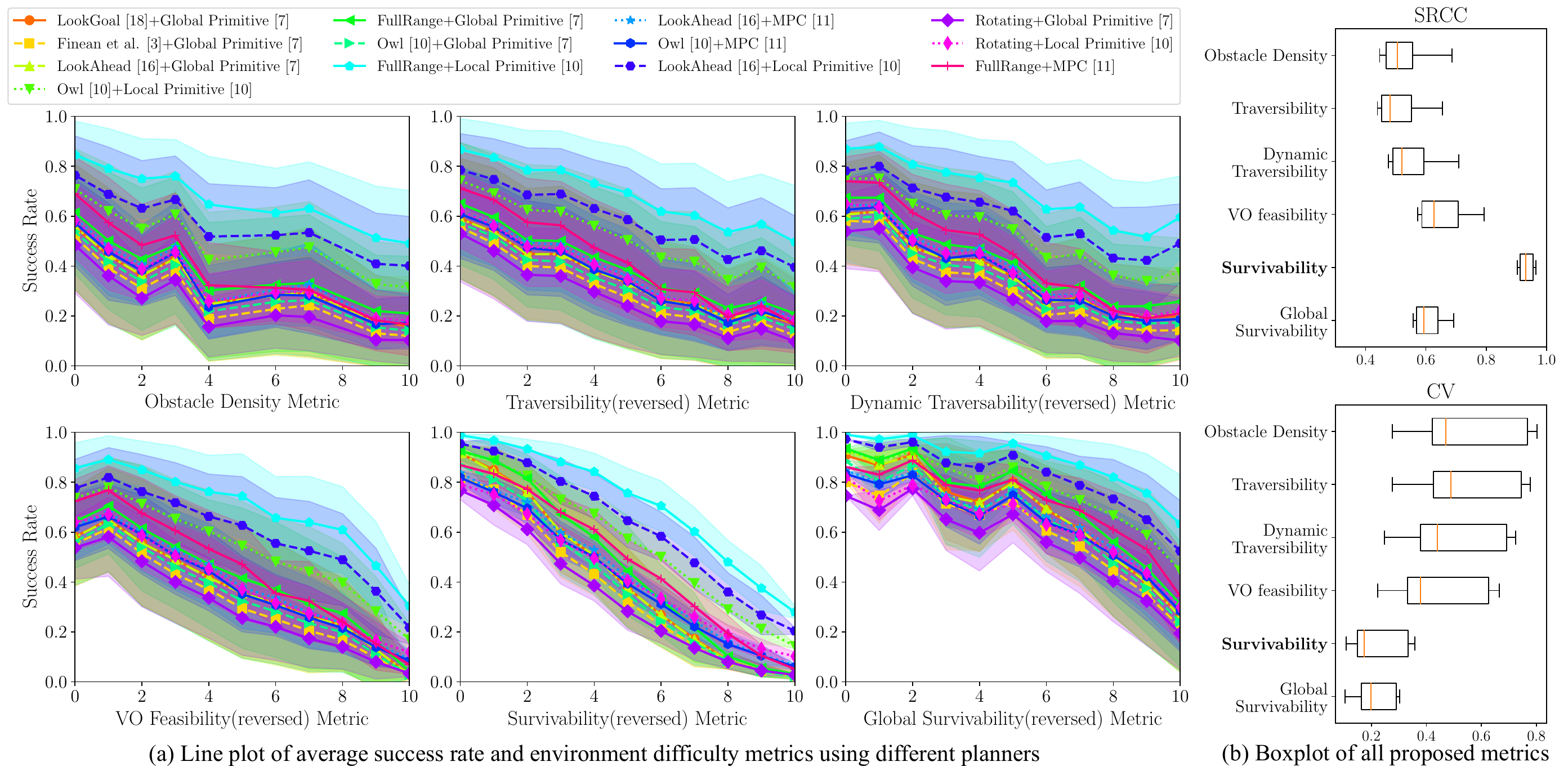}
    \caption{All experiment results of the proposed metrics. In (a), the plot shows the relationship between the success rate and the pre-processed metrics for different planners. Different colors denote planners. The light-colored band surrounding the curve represents one standard deviation of the success rate of each planner under each level of the difficulty metric. In (b), the boxplot shows SRCC and CV values of different metrics.}
    \label{fig: experiment_results_all}
\end{figure*}

\subsection{Experiment Results}\label{sec:experiment_results}
For each planner on each map, we experiment with multiple trials to calculate a corresponding success rate. In these trials, we apply different start and target positions and different robot velocities. There are 9 different position candidates for the start and target positions, and the robot velocity is chosen from $\{2,4,6\}$ m/s. Therefore, we have ${9\choose 2}  \times 3=216$ trials for each planner on each map to calculate the success rate. 

The metrics $D(M_i)$ of these maps undergo two pre-processing steps: normalization and reverse. Normalization scales the metric value so that $D(M_i)\in[0, 10]$. Reverse will let $D(M_i)=10-D(M_i)$ for some metrics to ensure that the high metric always indicates difficult maps as explained in Tab. \ref{tab:metric_correlation}. Then we show the correlation figures between the metrics and the success rate in Fig. \ref{fig: experiment_results_all}. The quantitative comparison of different metrics can also be seen in Tab. \ref{tab:result}. The survivability metric achieves the best SRCC and CV, and it can be demonstrated from the line plot in Fig. \ref{fig: experiment_results_all} (a) that the survivability shows a good monotonical relationship with the success rate of every planner.

\begin{table}[!h]
    \centering
    \begin{tabular}{@{}cccc@{}}
    \toprule
    \textbf{Difficulty Metric} & \textbf{SRCC} & \textbf{CV} & \textbf{Time} \\
    \midrule
    Obstacle Density & $0.501\pm 0.068$ & $0.571\pm 0.193$ & $\textbf{2s}$ \\
    Traversability & $0.511\pm 0.068$ & $0.568\pm 0.184$ & $7s$ \\
    Dynamic Traversability & $0.551\pm 0.073$ & $0.516\pm 0.173$ & $30s$ \\
    VO feasibility & $0.651\pm 0.077$ & $0.461\pm 0.161$ & $7s$ \\
    Survivability & $\textbf{0.932}\pm \textbf{0.023}$ & $\textbf{0.230}\pm\textbf{0.095}$ & $20s$ \\
    Global Survivability & $0.607\pm 0.044$ & $0.242\pm 0.078$ & $20s$ \\
    \bottomrule
    \end{tabular}
    \caption{Evaluation of metrics in terms of their SRCC, CV, and calculation time of the metric for one map.}
    \label{tab:result}
\end{table}

\section{DISCUSSION}
In this section, we further evaluate the difficulty metrics from sec. \ref{sec:metric} using SRCC and CV evaluation method introduced in section \ref{sec:eval_method}. Sec. \ref{sec:metric_scope} analyzes the reason why each metric works or not, and Sec. \ref{sec:use_case} presents the specific use case of the best metric Survivability.
\subsection{Scope of Metrics for Difficulty Evaluation}
\label{sec:metric_scope}

1) Obstacle density: Obstacle density mainly evaluates static environment difficulty, explaining its weak correlation with success rate in dynamic settings where velocity is not considered. We further validate this by grouping maps based on obstacle velocities and investigating the correlation in each group. Tab. \ref{tab: combined_obs_vel} shows that the correlation in each group improves significantly, as reflected by higher SRCC and lower CV values.
  
\begin{table}[!h]
    \centering
    \begin{tabular}{@{}lccc@{}}
    \toprule
    \textbf{Metric} & \textbf{Obstacle Velocity} & \textbf{SRCC}             & \textbf{CV}               \\ \midrule
    \multirow{4}{*}{Obstacle Density} & $2.0-6.0$m/s & $0.501\pm 0.068$ & $0.571\pm 0.193$ \\
    & $2.0$m/s & $0.788\pm 0.045$ & $0.075\pm 0.026$ \\
    & $4.0$m/s & $0.784\pm 0.030$ & $0.158\pm 0.050$ \\
    & $6.0$m/s & $0.758\pm 0.029$ & $0.272\pm 0.109$ \\ \midrule
    \multirow{4}{*}{VO Feasibility}   & $2.0-6.0$m/s & $0.651\pm 0.077$ & $0.461\pm 0.161$ \\
    & $2.0$m/s & $0.945\pm 0.017$ & $0.006\pm 0.002$ \\
    & $4.0$m/s & $0.953\pm 0.005$ & $0.124\pm 0.038$ \\
    & $6.0$m/s & $0.946\pm 0.007$ & $0.214\pm 0.079$ \\ \bottomrule
    \end{tabular}
    \caption{SRCC and CV of metrics after grouping the maps according to the obstacle velocity. The reported improvements are in comparison to their original values.}
    \label{tab: combined_obs_vel}
\end{table}

2) \& 3) Traversability and Dynamic Traversability: Similar to obstacle density, the poor performance of traversability and dynamic traversability can also be attributed to neglecting obstacle velocity.

4) VO Feasibility: Since different configurations of obstacle velocities can result in variations in VO feasibility regions, it is expected to be a good metric for dynamic environments. However, it does not have a good correlation with the success rate, featured by low SRCC and high CV. As demonstrated in Tab.  \ref{tab: combined_obs_vel}, the SRCC and CV values become significantly better if the obstacle velocities are the same. This indicates that the VO feasibility metric is similar to the metrics for static environments, which are not sensitive to the change in obstacle velocities. This limitation can be traced back to the generation process of VO infeasible areas. For two maps with obstacles at the same position but different velocities, the infeasible area only shifts directionally. The size of these infeasible regions remains unchanged. Thus, the VO feasibility metric cannot capture dynamic environment difficulty variations caused by different obstacle velocities.

5) Survivability: The Survivability metric consistently scores the highest in SRCC and lowest in CV, revealing a strong correlation between survivability and success rate, meaning the metric is effective in assessing dynamic map difficulty in \textit{Dataset I}. We also test survivability with higher sampling density. By reducing the distance between sample positions, $d_{sample}$ from 12.5 m to 10 m, the number of samples increases from 9 to 16 in each map. However, the SRCC only demonstrates a small improvement from 0.932 to 0.941 at the cost of 2 times the calculation time. 

We further investigate the generalization ability of survivability by fitting a Gaussian distribution to the success rate of \textit{Dataset I} under the same survivability metric and then calculating the Mahalanobis Distance of the data points in \textit{Dataset II} to the fitted Gaussian distribution. The average Mahalanobis Distance is 0.74, and 99.3\% of the data points in \textit{Dataset II} lie within the $3\sigma$ range of the fitted distribution. This suggests the survivability metric effectively evaluates the dynamic map difficulty when the sizes and velocities of obstacles differ in the map and when the motion profile is RVO. For example, if there are two maps with CVM and RVO motion profiles that have similar survivability, the success rate of one planner in these two maps is expected to be similar. The detailed analysis, including maps with different motion profiles, will be presented in the supplementary file.

6) Global Survivability: It does not correlate better with success rates than the original Survivability metric. Fig.  \ref{fig: experiment_results_all} shows that the variance of success rates for maps with low difficulty is small, while the variance of success rates for maps with higher difficulty is large. Since we calculate the minimum survival time of all sampled positions, most maps have similarly small Global Survivability and are thus considered difficult. It is hard to tell the difference between them by Global Survivability.

\subsection{Use Cases of Survivability Metric}
\label{sec:use_case}
There are three use cases of the survivability metric:
\subsubsection{Comparison of Different Planners}
Fig.  \ref{fig:compare} shows success rates for three trajectory planners with varied survivability levels. The \textit{Local Primitive} planner consistently outperforms others. While \textit{Global Primitive} starts stronger than \textit{MPC}, it is surpassed by \textit{MPC} as the difficulty increases. The conclusions here are based on the assumptions mentioned when we design the simulator in Sec. \ref{sec:assumption}.

Fig.  \ref{fig:compare} also compares five gaze planners and \textit{FullRange} Perception against survivability levels. Here, \textit{FullRange} sets the upper bound. \textit{LookAhead} and \textit{LookGoal} are top performers, with \textit{Rotating} lagging behind.

\begin{figure}[t!]
    \centering
    \includegraphics[width=0.48\textwidth]{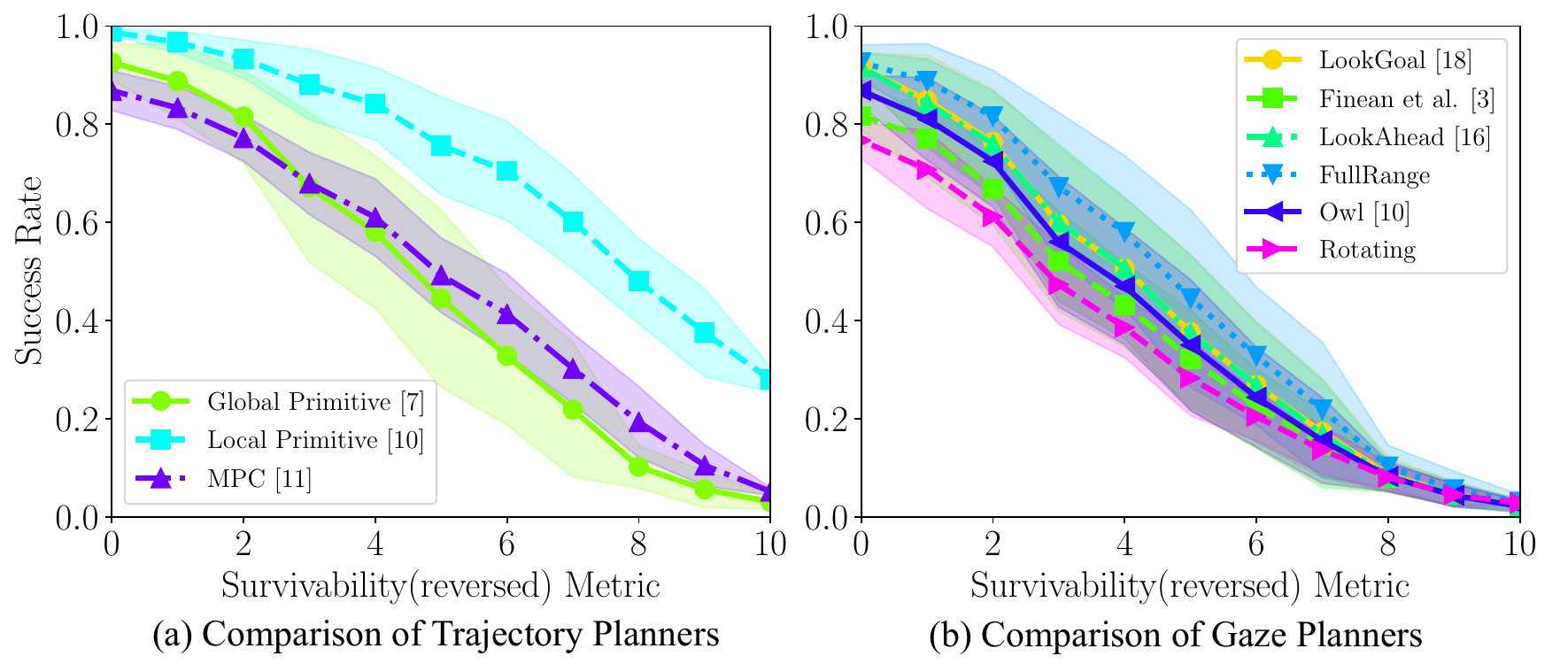}
    \caption{The comparison of planners using the survivability metric.}
    \label{fig:compare}
\end{figure}

\subsubsection{Generate Maps with Predefined Survivability Metric}
For benchmarking collision avoidance methods, it is also important to generate maps with pre-defined difficulty so that we can gradually test our method in harder scenarios. In \textit{Dataset I}, the survivability of a map is determined by: $n_{obs}$, $r_{obs}$, $v_{obs}$. We use a linear regression model to fit this relationship:
\begin{equation}
    S = f(n_{obs}, r_{obs}, v_{obs}) = \beta_0 + \beta_1 n_{obs} + \beta_2 r_{obs} + \beta_3 v_{obs}
\end{equation}

The resulting coefficients are -6.014, 0.226, 2.646, and 1.104 for $\beta_0$, $\beta_1$, $\beta_2$, and $\beta_3$, respectively. Using the fitted model $f$, we can generate maps with a specified survivability metric $S$. The process is formulated as an Integer Linear Programming (ILP) optimization problem:
\begin{align}
    \text{minimize} \quad & \left|f(n_{obs}, r_{obs}, v_{obs}) - S\right| \\
    \text{s.t. } \quad & n_{obs} \in \mathbb{Z}, 10 \leq n_{obs} \leq 30, \\
    & 0.5 \leq r_{obs} \leq 1.5, 2 \leq v_{obs} \leq 6 
\end{align}

\subsubsection{Calculating Survivability Metric in Other Simulator and real world}
We can easily calculate survivability in other high-fidelity simulators. An example using the gazebo simulation is shown in Fig. \ref{fig: metric_gazebo}. For the real-world test, survivability can also be calculated by recording the obstacles as mentioned in Sec. \ref{sec:survivability}.
\begin{figure}[t!]
    \centering
    \begin{subfigure}[t]{0.465\linewidth}
        \centering
        \includegraphics[width=0.99\textwidth]{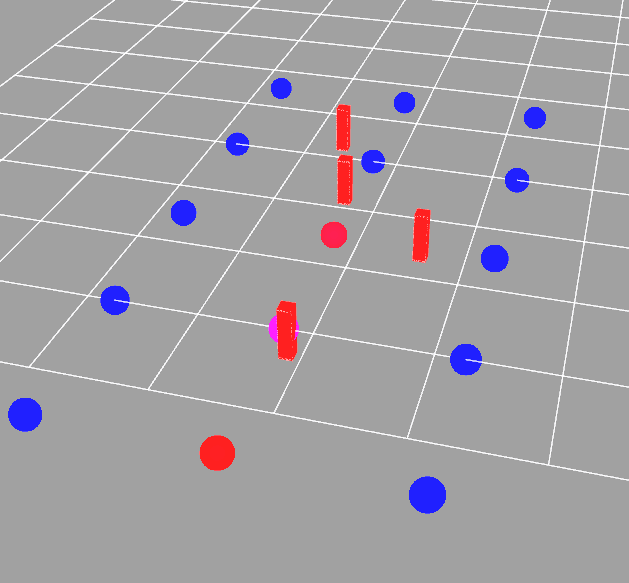}
        \caption{Metric Calculation}
    \end{subfigure}
    \begin{subfigure}[t]{0.45\linewidth}
        \centering
        \includegraphics[width=0.99\textwidth]{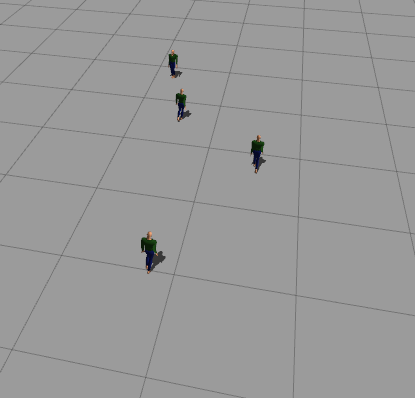}
        \caption{Gazebo Simulation}
    \end{subfigure}
    \caption{The calculation of the survivability metric in a gazebo simulation environment. In (a), the pillars represent the moving pedestrians. The spheres represent the sampled robot positions, where red ones have already collided with obstacles and blue ones have not.}
    \label{fig: metric_gazebo}
\end{figure}

\section{CONCLUSIONS}
In this paper, we propose four metrics to evaluate the environmental difficulty of dynamic environments for collision avoidance problems. We validate their effectiveness through extensive experiments on our custom simulator and provide a detailed analysis of the results, aiming to demonstrate insights into the limitations of these metrics and their potential applications. Results show that the proposed survivability metric is suitable for assessing the dynamic environment difficulty. VO Feasibility metrics can also be used for rapid evaluations in consistent obstacle velocities. Our future work will explore the applicability of these metrics in 3D scenarios to further expand their scope of usage.











\bibliographystyle{IEEEtran}

\bibliography{ref}

\end{document}